\documentclass{article}

\usepackage[preprint]{neurips_2025}

\usepackage[utf8]{inputenc}
\usepackage[T1]{fontenc}
\usepackage{amsfonts}
\usepackage{amsmath}
\usepackage{amssymb}
\usepackage{booktabs}
\usepackage{graphicx}
\usepackage{subfig}
\usepackage{nicefrac}
\usepackage{microtype}
\usepackage{url}
\usepackage{xcolor}
\usepackage{makecell}
\usepackage{multirow}
\usepackage{caption}
\usepackage{tabularx}
\usepackage{colortbl}
\usepackage{pgfplots}
\usepackage{stackengine}
\usepackage{csquotes}
\usepackage{enumitem}
\usepackage{hyperref}
\hypersetup{
    colorlinks=true,
    linkcolor=blue,
    filecolor=blue,
    urlcolor=blue,
    citecolor=blue,
}

\pgfplotsset{compat=1.17}
\newcommand{\cellcolorgrey}[1]{ 
    \begingroup
    \edef\tempa{#1}
    \ifx\tempa\empty%
    \else
        \pgfmathsetmacro{\scaled}{max(0,min(100,round(100*abs(#1))))}
        \xdef\clr{gray!\scaled!white}
        \cellcolor{\clr}{#1}
    \fi
    \endgroup
}

\graphicspath{{figs/}}
\makeatletter
\def\input@path{{figs/}}
\makeatother

\title{Predicting Privacy Leakage from Weight Spectral Density}

\author{%
  Richard~J.~Preen\\
  Department of Computer Science and Creative Technologies\\
  University of the West of England\\
  Bristol, UK BS16 1QY\\
  \texttt{richard2.preen@uwe.ac.uk}\\
  \And
  Jim~Smith\\
  Department of Computer Science and Creative Technologies\\
  University of the West of England\\
  Bristol, UK BS16 1QY\\
  \texttt{james.smith@uwe.ac.uk}\\
}

\begin{document}

\maketitle

\begin{abstract}
Membership inference attacks (MIAs) are widely used to audit the privacy disclosure risk of machine learning models, however current state-of-the-art attacks require training computationally expensive shadow models, making large-scale privacy evaluation impractical. In this work, we investigate whether inexpensive spectral metrics derived from the heavy-tailed self-regularisation framework can serve as proxies for MIA vulnerability. We evaluate several WeightWatcher spectral metrics on image and tabular classification tasks and compare their relationship with MIA privacy leakage against conventional measures of generalisation. Across datasets, stable rank exhibits a strong positive correlation with overall MIA success, while Log $\alpha$-Norm shows a consistent negative correlation with MIA vulnerability at the low false-positive regime. These associations are observed to be stronger than those obtained using the generalisation gap. The results indicate that neural network spectra may contain information about privacy leakage that is not fully captured by conventional measures of overfitting, motivating spectral analysis as a promising direction for scalable privacy auditing.
\end{abstract}

\section{Introduction}%
\label{sec:introduction}

Machine learning models are increasingly trained on sensitive data held within trusted research environments (TREs)\footnote{Also known as safe havens, secure data environments, and secure research services.}. Before such a model can be released from a TRE, its owners are required to demonstrate that it does not disclose sensitive information about the individuals in its training set~\citep{Jefferson:2022}. Membership inference attacks (MIAs)~\citep{Shokri:2017} have become one of the standard methods used to quantify the overall risk as part of the auditing process. An MIA attempts to determine whether a specific record was used to train a target model, and tools such as SACRO-ML~\citep{Smith:2025} now package them for use directly within TRE disclosure-control workflows.

The most powerful of the current state-of-the-art MIAs, such as the likelihood ratio attack (LiRA)~\citep{Carlini:2022} and the robust membership inference attack (RMIA)~\citep{Zarifzadeh:2024}, typically rely on training reference (shadow) models to approximate the target model's behaviour. This makes them computationally expensive, and in many settings prohibitively so. The problem is compounded in TRE auditing contexts where many candidate models may need to be screened before release.

A parallel line of research has studied the phenomenon of grokking~\citep{Power:2022}, wherein models undergo a delayed transition from memorisation to generalisation long after training loss has converged. This observation raises a broader question: what internal properties of a trained neural network's weight structure distinguish models that memorise from those that generalise, and can such properties be measured cheaply, without access to the data used to fit the model?

WeightWatcher (WW)~\citep{Martin:2021a,Martin:2021b,Prakash:2025} offers a promising lens through which to examine this question. Grounded in the theory of heavy-tailed self-regularisation (HT-SR), WW characterises the empirical spectral density (ESD) of a model's weight matrices from the trained weights alone, making them cheap-to-compute. These metrics have been linked empirically to a model's degree of implicit self-regularisation and, in turn, to its generalisation performance.

This raises the question: if WW metrics reflect the degree to which a model has generalised versus memorised its training data, might they also serve as proxies for privacy leakage? Intuitively, a model whose weight matrices remain close to random has not organised its parameters around training examples, whereas a model exhibiting strong power-law structure may have done so in ways that are recoverable by an attacker. Yet the relationship between spectral properties and susceptibility to MIA is far from obvious, and no prior work has investigated it empirically.

In this paper, we investigate whether WW metrics correlate with generalisation gap and vulnerability to MIAs such as LiRA. If such correlations exist, WW could provide a computationally tractable, data-free method for estimating privacy risk, becoming a powerful tool for practitioners who cannot afford the overhead of shadow model pipelines.

In particular, this paper makes the following contributions:

\begin{itemize}
    \item We conduct the first empirical study of the relationship between neural network weight spectral metrics and vulnerability to MIA.
    \item We show that WW stable rank is strongly associated with overall MIA success, and that this association is consistently stronger than that of the generalisation gap.
    \item We show that WW Log $\alpha$-Norm is consistently associated with MIA success in the low false-positive-rate regime, where the generalisation gap is a weak and inconsistent predictor.
    \item We show that spectral metrics and the generalisation gap capture complementary information about privacy leakage, and that combining them improves prediction of MIA vulnerability over either signal alone.
    \item We demonstrate that spectral metrics can be used to identify high-risk models as a binary classification task, indicating practical utility for scalable, data-free privacy auditing.
\end{itemize}

\section{Background}%
\label{sec:background}

\subsection{Heavy-Tailed Self-Regularisation and WeightWatcher}
\label{sec:htsr}

WW is an open source\footnote{\url{https://github.com/CalculatedContent/WeightWatcher}}  diagnostic tool for analysing trained deep neural networks without requiring access to training or test data~\citep{Martin:2021a,Martin:2021b}. It is based on the theory of HT-SR, which analyses the ESD of the correlation matrix associated with each layer's weight matrix. The central finding of HT-SR theory is that well-trained, generalisable networks are not random: rather than resembling the spectra of random matrices, their layer-wise ESDs display structured, self-organised correlations that emerge during optimisation, in many cases converging to heavy-tailed, power-law form~\citep{Martin:2021a}. This behaviour is interpreted as a signature of implicit self-regularisation, analogous to the strong correlations observed in self-organising physical systems, and has been proposed as an explanation for why overparameterised networks generalise well in practice.

\cite{Martin:2021a} characterise this process in terms of six qualitative spectral phases through which a layer can progress as training proceeds: (i) \emph{random-like}, in which the weights remain statistically indistinguishable from a random matrix; (ii) \emph{bleeding-out}, in which small deviations from randomness first appear; (iii) \emph{bulk+spikes}, in which discrete signal components emerge above an otherwise random bulk; (iv) \emph{bulk-decay}, in which the bulk itself begins to decay as correlations strengthen; (v) \emph{heavy-tailed}, in which scale-free correlations dominate the spectrum and self-regularisation is strong; and (vi) \emph{rank-collapse}, in which over-regularisation causes most of the layer's effective information to be lost. Hyperparameters that govern the implicit regularisation strength of stochastic gradient descent (SGD), most notably batch size, have been shown to move models through these phases, with smaller batch sizes typically producing stronger implicit self-regularisation.

To characterise where a layer sits along this progression, WW fits the tail of each layer's ESD to a (truncated) power law and reports a small set of summary statistics. The power-law exponent $\alpha$ describes the shape of this tail; under HT-SR theory, models with $\alpha$ in the range $[2, 4]$ are considered well regularised, whereas $\alpha < 2$ is associated with over-fitting and rank collapse and $\alpha > 4$ with an under-trained, near-random layer. Norm-based metrics such as the log $\alpha$-norm and the log spectral norm combine this shape information with the scale of the spectrum, making them more directly comparable across layers and architectures of different width and depth. A further metric, stable rank, measures the effective dimensionality of a weight matrix and is a noise-tolerant alternative to the rank of the weight matrix. Because all of these quantities are computed directly from trained weights, WW metrics have been proposed as scalable proxies for generalisation performance that require no forward passes over data~\citep{Martin:2021b,Prakash:2025}, motivating our investigation of whether they might similarly serve as proxies for a model's susceptibility to membership inference.

\subsection{Membership Inference Attacks}

MIAs aim to determine whether a specific record was used to train a target model, and have become a standard tool for empirically auditing privacy disclosure~\citep{Shokri:2017}. Broadly, existing attacks trade off attack strength against computational cost: the strongest attacks require training many reference (shadow) models of similar architecture and under the same conditions as the target, while a growing body of work seeks to approximate this signal more cheaply.

\subsubsection{Reference-model attacks}

LiRA~\citep{Carlini:2022} is widely regarded as the closest practical approximation to a record's ground-truth vulnerability. LiRA trains multiple shadow models both with and without the target record, fits Gaussian models to the resulting IN and OUT loss distributions, and performs a likelihood-ratio test to infer membership. Because it directly approximates the leave-one-out counterfactual behaviour of the target model, LiRA achieves strong, well-calibrated performance across architectures and datasets, particularly on rare or outlier records. Its principal drawback is cost: obtaining reliable estimates typically requires dozens of shadow models trained under matched conditions, which is prohibitive for large modern architectures or datasets.

Subsequent work has sought to reduce this overhead while retaining LiRA's discriminative power. RMIA~\citep{Zarifzadeh:2024} reformulates membership inference as a pairwise test in which a target sample is compared against many population samples, using a small number of reference models together with a large pool of unlabelled population data to normalise the resulting likelihood ratio; this substantially reduces the number of reference models required relative to LiRA while achieving comparable power. \cite{Galichin:2025} instead reduce the cost of training the shadow models themselves by incorporating knowledge distillation from the target model, aligning shadow model behaviour with the target more closely and reporting improvements over LiRA and related loss-trajectory attacks on image classification benchmarks, though training remains the dominant cost. Across these variants, the requirement to retrain one or more models under conditions approximating the target remains the primary obstacle to using reference-model attacks for routine, large-scale privacy auditing.

\subsubsection{Reference-free and one-shot estimators}

A parallel line of work avoids reference-model training entirely, instead exploiting signals available from the target model alone. White-box attacks that inspect gradients or intermediate activations have shown that even well-generalised models can leak substantial membership information: \cite{Nasr:2019} show that gradients, particularly from later layers, are more informative than either activations or outputs, and that this leakage is not well predicted by a model's generalisation gap. \cite{Li:2024} similarly use statistical neuron-selection methods together with a single shadow model to identify which neurons carry the most membership-relevant signal, while related work on interpretability-based attacks shows that feature-importance information can itself be exploited to distinguish members from non-members~\citep{Liu:2024}.

Fully black-box, shadow-free alternatives have also been proposed: \cite{Liu:2023} infer membership from the Jacobian norm of the target model's predictions, clustering samples by sensitivity, and can operate from as little as a single query record. One-shot, label-only attacks exploit the relative robustness of member samples to adversarial perturbation~\citep{Peng:2024}, while the GLiR attack of \cite{Leemann:2023} performs one-shot auditing from training-time gradient statistics without any shadow models, and \cite{Suri:2024} show that white-box access to inverse-Hessian information can outperform black-box loss-based attacks under SGD, albeit at considerable computational cost and with access to the full training set. \cite{Steinke:2023} and \cite{Andrew:2024} take a complementary approach, deriving privacy-auditing estimates from a single training run by treating a random subset of the training data as an implicit control group, avoiding the need to train separate reference models altogether.

Other estimators dispense with reference models entirely, avoiding the cost of retraining, though most still rely on some form of auxiliary data or held-out computation. QMIA~\citep{Bertran:2023} trains a quantile regression model on held-out data to estimate a per-record confidence threshold as a function of the input itself, conditioning thresholds on the local difficulty of a record rather than using a single global threshold as in earlier loss-based attacks~\citep{Yeom:2018}; a related approach has been used to detect training-set membership for documents used to train large language models~\citep{Zhang:2024}. QMIA requires no reference models and scales well, but may perform poorly in sparse regions of the input distribution where few comparable records are available for quantile estimation, precisely where disclosure risk is often greatest.

Training-dynamics estimators such as LT-IQR~\citep{Pollock:2025} instead exploit the trajectory of a record's loss over training, treating records whose loss falls unusually far as more likely to have been memorised; such methods can provide useful relative rankings when full training trajectories are retained, but typically stop short of calibrated disclosure probabilities. Most closely related to our approach, \cite{Dodd:2025} propose a model-level, reference-free estimator based on the observation that memorisation suppresses the heavy-loss tail of the training-loss distribution, and estimate aggregate disclosure from the separation between training and test-loss tails without any additional training. Like WW metrics, this approach requires no reference models and is computationally cheap, but provides only an aggregate, model-level signal rather than per-record vulnerability estimates.

\subsubsection{Cheap-to-compute privacy proxy metrics}

Because MIAs succeed by exploiting differences in a model's behaviour on training versus held-out data, a model's generalisation gap has long been used as an inexpensive, if coarse, proxy for its privacy risk~\citep{Yeom:2018,Shokri:2017}. Larger gaps between training and test performance are generally associated with greater vulnerability to MIAs, and the generalisation gap requires no additional model training beyond the target model itself. However, this relationship is not always reliable: \cite{Nasr:2019} find that generalisation error is a poor predictor of privacy risk in white-box settings, since large, expressive models may memorise individual training records in ways that are not reflected in aggregate held-out performance. This gap between an easily-measured, coarse-grained statistic and a model's true susceptibility to attack is a central motivation for identifying alternative, equally inexpensive metrics that more directly track privacy leakage.

Drawing on traditional statistical disclosure control concepts, including degrees of freedom and $k$-anonymity, \cite{Preen:2026} investigate the use of cheap-to-compute model-level metrics to reduce the computational cost of MIA assessment. They show that for tree-based classification models, structural measures derived from trained models and their predictions provide high-precision but low-recall indicators of MIA vulnerability. This enables high risk models to be identified without needing to perform expensive reference model-based MIA assessment. These results demonstrate the potential for inexpensive metrics to complement more intensive MIA assessment, reducing the overall computational cost of privacy auditing in secure data facilities.

\subsection{Summary}

Taken together, this literature reveals a persistent trade-off: reference-model attacks such as LiRA and RMIA offer the most reliable estimates of privacy leakage but are computationally prohibitive to apply broadly, while existing reference-free alternatives either require additional held-out data and model training (QMIA), depend on retaining full training trajectories (LT-IQR), or provide only coarse, aggregate disclosure estimates (loss-tail methods). The generalisation gap remains the most widely used training-free proxy, yet its relationship with MIA vulnerability is inconsistent, and it can be a noisy predictor when models diverge primarily in how much they memorise their training data rather than in their held-out performance. WW metrics are, by construction, even cheaper to obtain than the generalisation gap, requiring no evaluation data at all, and are grounded in a theory that explicitly links a model's weight-space structure to the degree of implicit regularisation, and hence to overfitting and memorisation. However, no prior work has directly examined whether these spectral metrics track vulnerability to state-of-the-art MIAs such as LiRA, nor whether they carry information about privacy leakage beyond that captured by the generalisation gap.

\section{Methodology}%
\label{sec:methodology}

To study the relationship between WW measures of generalisation and MIA risk, we train a range of deep neural networks with varying depths and widths and with various hyperparameter configurations. For each of these models, we measure the MIA risk using the open source\footnote{\url{https://github.com/AI-SDC/SACRO-ML}} SACRO-ML~\citep{Smith:2025} implementation of LiRA~\citep{Carlini:2022}. In addition, we report WW and target model metrics, including generalisation gap, and model accuracy.

\subsection{Metrics}

We evaluate models along three axes: generalisation, spectral complexity via WW, and privacy leakage via LiRA. These metrics are summarised in Table~\ref{tab:metrics} and further detailed below.

\begin{table}[t]
    \centering
    \caption{Summary of evaluation metrics. WW metrics are computed directly from model weights without access to data; target model and MIA metrics require data and/or shadow models.}
    \label{tab:metrics}
    \small
    \begin{tabular}{llll}
        \toprule
        \textbf{Category} & \textbf{Metric} & \textbf{Description}\\
        \midrule
        \multirow{3}{*}{Generalisation}
        & Train accuracy & Fit to training data\\
        & Test accuracy & Fit to held-out data\\
        & Generalisation gap & $\epsilon_{\text{gap}} = \epsilon_{\text{test}} - \epsilon_{\text{train}}$\\
        \midrule
        \multirow{4}{*}{\shortstack[l]{WW (spectral)}}
        & $\alpha$ & Power-law tail exponent of ESD; shape metric\\
        & Log $\alpha$-norm & Norm-based shape and scale metric\\
        & Log spectral norm & Log max singular value\\
        & Stable rank & Effective dimensionality of weight matrix\\
        \midrule
        \multirow{2}{*}{Privacy (LiRA)} & AUC & Overall MIA discriminative power\\
        & TPR@0.001 & Attack success at low false-positive rate\\
        \bottomrule
    \end{tabular}
\end{table}

\paragraph{Generalisation metrics.}
We record train accuracy, test accuracy, and generalisation gap, defined as $\epsilon_{\text{gap}} = \epsilon_{\text{test}} - \epsilon_{\text{train}}$. A large generalisation gap indicates that the model has memorised training examples rather than learned a broadly applicable function, and is therefore expected to be more susceptible to MIA~\citep{Shokri:2017}.

\paragraph{Spectral metrics.}
WW analyses the ESD of each layer's correlation matrix $\mathbf{X} = \mathbf{W}^\top \mathbf{W}$, where $\mathbf{W}$ is the layer weight matrix, without requiring access to any training or test data. For each layer, it fits the tail of the ESD to a (truncated) power-law distribution and extracts summary statistics that characterise the shape and scale of the spectrum. We use the following layer-aggregated metrics:

\begin{itemize}

    \item $\boldsymbol{\alpha}$ \textbf{(Power-law exponent).}
    The (negative) slope of the tail of the ESD on a log-log scale. Under HT-SR theory, smaller $\alpha$ indicates stronger implicit self-regularisation. The range $\alpha \in [2, 4]$ is considered healthy; $\alpha < 2$ suggests overfitting and rank collapse, while $\alpha > 4$ indicates an under-trained or near-random layer.

    \item \textbf{Log $\alpha$-norm.}
    A norm-based metric that accounts for both shape and scale and is suitable for comparing networks with differing hyperparameters and depths simultaneously.

    \item \textbf{Log spectral norm.}
    The logarithm of the largest singular value of $\mathbf{W}$, which is useful to compare models of different depths at a coarse grain level.

    \item \textbf{Stable rank.}
    Defined as $\mathcal{R}(\mathbf{W}) = \|\mathbf{W}\|_F^2 / \|\mathbf{W}\|_2^2$, the stable rank measures the effective dimensionality of the weight matrix. It is a robust, noise-tolerant alternative to the matrix rank. Higher stable rank indicates that the layer remains more random-like with weaker implicit self-regularisation. In HT-SR theory, this lack of regularisation provides the model with excess capacity, suggesting a greater tendency for memorisation.

\end{itemize}

\paragraph{Membership inference metrics.}
We evaluate privacy leakage using the online variant of LiRA~\citep{Carlini:2022}, which frames membership inference as a likelihood ratio test over the outputs of shadow models; here we use 64 shadow models. We report two standard metrics: (i)~\textbf{AUC}, the area under the ROC curve, where $0.5$ corresponds to a random-guess attacker and $1.0$ to a perfect attacker; and (ii)~\textbf{TPR@0.001}, the true positive rate at a fixed false positive rate of $0.001$, which measures attack success in the low-false-positive regime most relevant to realistic adversarial settings~\citep{Carlini:2022}.

\subsection{Datasets and Target Models}

\textbf{Datasets.} For initial exploration we use CIFAR-10, a 10-class image classification dataset with 60,000 $32\times32\times3$ RGB images; and the OpenML\footnote{\url{https://www.openml.org}} Volkert dataset (ID: 41166), a 10-class tabular dataset with 58,310 instances and 180 numeric features. For the Volkert dataset, features are normalised with zero mean and unit variance.

\textbf{Architectures.} All models are feedforward multi-layer perceptrons (MLPs) with ReLU activations, no dropout or normalisation layers are used. A variety of depths $D$ (number of hidden layers) and widths $W$ (units per hidden layer), denoted $D \times W$ are used to include a range of target models. All architectures contain a 10-class linear output layer. Cross-entropy loss with implicit softmax is used for training.

\textbf{Training configurations.} For each architecture, we train models across a grid of hyperparameters: learning rates $\in \{0.001, 0.01\}$ and weight decay $\in \{0, 0.0005\}$. Stochastic gradient descent with a momentum of 0.9 and batch size of 32 is used for optimisation. All models are initialised using Kaiming uniform initialisation~\citep{He:2015}. To capture varying levels of memorisation, we evaluate models after 100 epochs. This yields $11 \text{ architectures} \times 2 \text{ learning rates} \times 2 \text{ weight decay values} = 44$ distinct target models with diverse training trajectories and generalisation properties.

While it has become standard practice to use a 50-50\% train/test split to compare different MIAs~\citep{Carlini:2022}, here we use the standard 50,000/10,000 split for CIFAR-10 and a random stratified 80-20\% split for the Volkert dataset to reflect a more realistic data availability when auditing real-world models.

Table~\ref{tab:hyperparams} summarises the architectures (network type) and hyperparameters (learning rate, epochs, and weight decay). Other components (such as activations) remain fixed across all experiments to isolate the effects of the varied hyperparameters.

\begin{table}[t]
    \centering
    \caption{Target model architectures and hyperparameters; 44 total models.}%
    \label{tab:hyperparams}
    \small
    \begin{tabular}{llll}
        \toprule
        \textbf{Category} & \textbf{Hyperparameter} & \textbf{Values} & \textbf{Rationale / Notes} \\
        \midrule
        \multirow{3}{*}{Architecture}
        & Network type & \makecell[l]{MLP-$D{\times}W$\\
        for $(D,W) \in \{(1,4096),$\\
        $(2,1024), (2,2048),$\\
        $(3,512), (3,1024),$\\
        $(3,2048), (4,1024),$\\
        $(4,2048), (4,4096),$\\
        $(8,512), (8,1024)\}$} &
        Varied depth $D$ and width $W$. \\
        & Activation & ReLU & Fixed across all runs. \\
        & Dropout / normalisation & None & Avoids confound with weight decay. \\
        \midrule
        \multirow{3}{*}{Optimisation}
        & Optimiser & SGD (momentum = 0.9) & Fixed across all runs. \\
        & Learning rate & \{0.001, 0.01\} & Low vs.\ high, implicit regularisation. \\
        & Weight decay & \{0, $5\times 10^{-4}$\} & Explicit $L_2$ regularisation strength. \\
        \midrule
        \multirow{3}{*}{Data}
        & Dataset fraction & 100\% & Fixed across all runs. \\ 
        & Data augmentation & None & Fixed across all runs. \\
        & Label noise & None (0\%) & Fixed across all runs. \\
        \midrule
        \multirow{2}{*}{Training}
        & Batch size & 32 & Fixed across all runs. \\
        & Epochs & 100 & Range of under and overfitting. \\
        \bottomrule
    \end{tabular}
\end{table}

\section{Results}%
\label{sec:results}

\begin{figure}
    \subfloat[]{\includegraphics[width=0.24\linewidth]{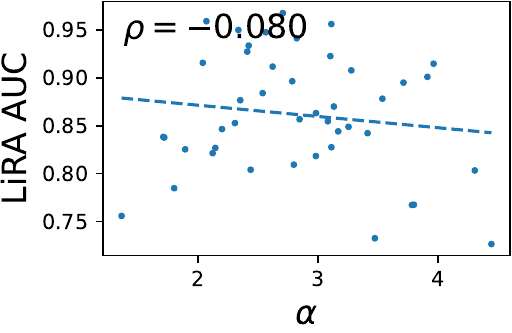}}
    \hfill
    \subfloat[]{\includegraphics[width=0.24\linewidth]{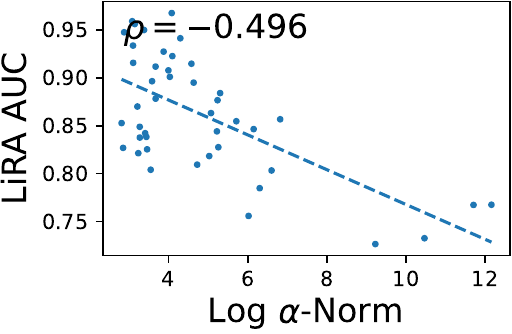}}
    \hfill
    \subfloat[]{\includegraphics[width=0.24\linewidth]{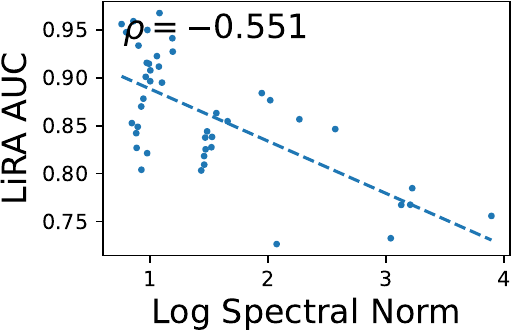}}
    \hfill
    \subfloat[]{\includegraphics[width=0.24\linewidth]{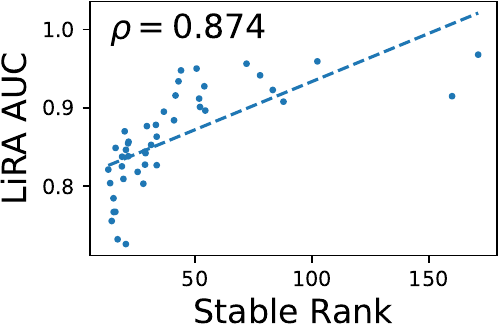}}\\
    \hfill

    \subfloat[]{\includegraphics[width=0.24\linewidth]{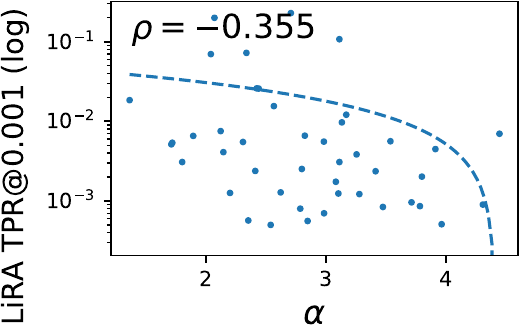}}
    \hfill
    \subfloat[]{\includegraphics[width=0.24\linewidth]{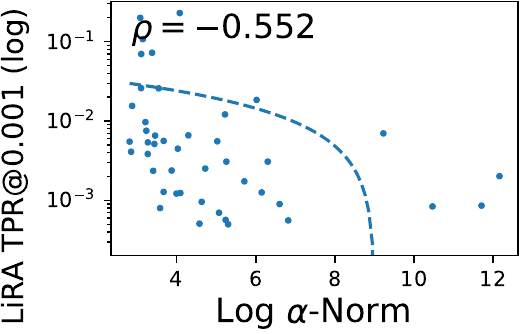}}
    \hfill
    \subfloat[]{\includegraphics[width=0.24\linewidth]{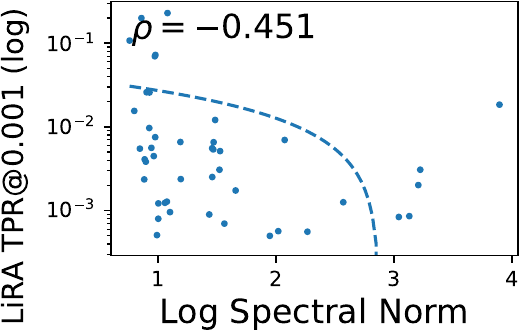}}
    \hfill
    \subfloat[]{\includegraphics[width=0.24\linewidth]{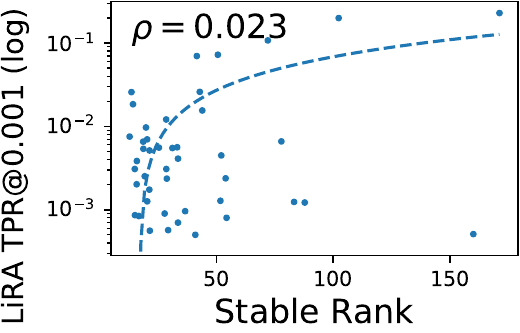}}\\
    \hfill

    \caption{CIFAR-10: Relationship between privacy risk (LiRA AUC and TPR@0.001) and spectral metrics ($\alpha$, log $\alpha$-norm, log spectral norm, and stable rank). LiRA TPR is shown on log-scale.}
    \label{fig:lira_ww_cifar}
\end{figure}

\begin{figure}
    \subfloat[]{\includegraphics[width=0.24\linewidth]{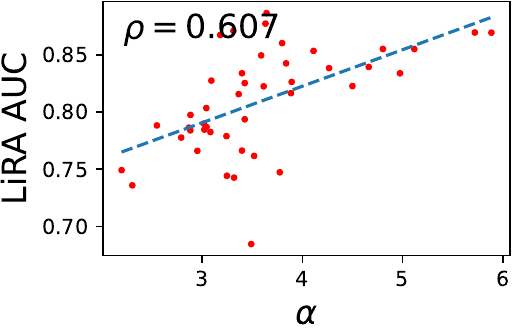}}
    \hfill
    \subfloat[]{\includegraphics[width=0.24\linewidth]{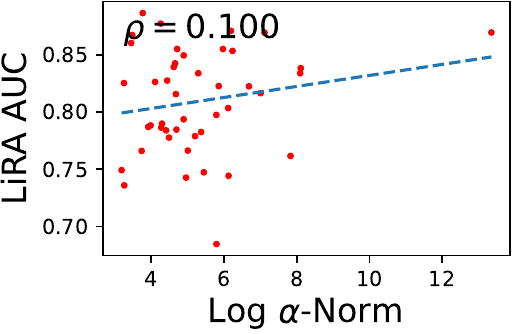}}
    \hfill
    \subfloat[]{\includegraphics[width=0.24\linewidth]{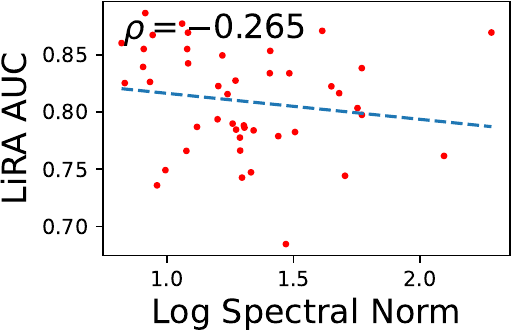}}
    \hfill
    \subfloat[]{\includegraphics[width=0.24\linewidth]{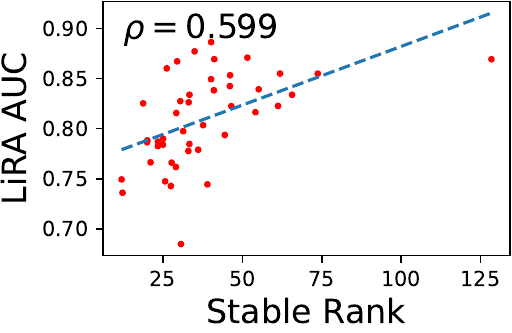}}\\
    \hfill

    \subfloat[]{\includegraphics[width=0.24\linewidth]{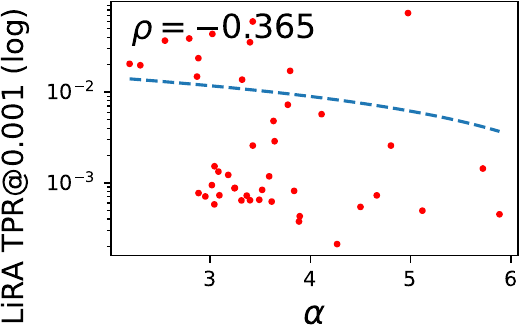}}
    \hfill
    \subfloat[]{\includegraphics[width=0.24\linewidth]{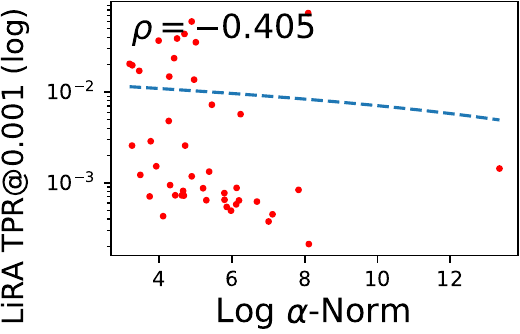}}
    \hfill
    \subfloat[]{\includegraphics[width=0.24\linewidth]{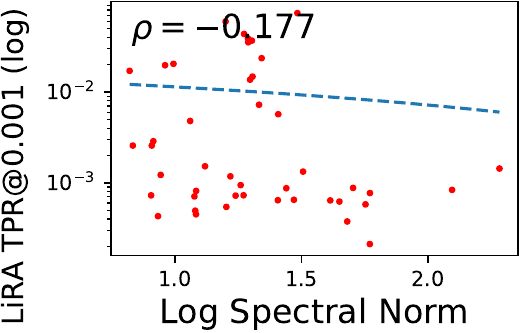}}
    \hfill
    \subfloat[]{\includegraphics[width=0.24\linewidth]{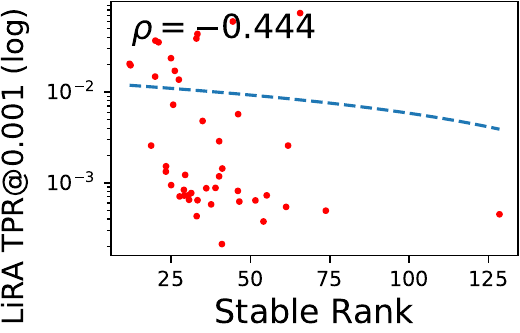}}\\
    \hfill

    \caption{Volkert: Relationship between privacy risk (LiRA AUC and TPR@0.001) and spectral metrics ($\alpha$, log $\alpha$-norm, log spectral norm, and stable rank). LiRA TPR is shown on log-scale.}%
    \label{fig:lira_ww_volkert}
\end{figure}

Figures~\ref{fig:lira_ww_cifar} and~\ref{fig:lira_ww_volkert} examine the relationship between spectral metrics and LiRA privacy leakage on the CIFAR-10 and Volkert datasets. Across both datasets, stable rank exhibits the strongest association with LiRA AUC, with Spearman's rank correlations of $\rho=0.87$ on CIFAR-10 and $\rho=0.60$ on Volkert ($p\le0.01$). In addition, Log $\alpha$-Norm is most strongly associated with LiRA TPR@0.001, with moderate negative correlations of $\rho=-0.55$ and $\rho=-0.40$ ($p\le0.01$). These results suggest that different spectral characteristics capture complementary aspects of privacy leakage, with stable rank reflecting overall attack success and Log $\alpha$-Norm better tracking leakage in the low FPR regime.

\paragraph{Relationship between spectral metrics and privacy leakage.}
Within the HT-SR framework, the power-law exponent $\alpha$ provides a measure of implicit regularisation, with $\alpha\approx2$ corresponding to well-regularised models and $\alpha\ll2$ indicating rank collapse. Here the relationship between $\alpha$ and LiRA AUC is seen to be inconsistent across datasets, with almost no correlation on CIFAR-10 ($\rho=-0.08$) but a moderate positive correlation on Volkert ($\rho=0.61$). However, $\alpha$ is consistently negatively correlated with LiRA TPR@0.001 across datasets ($\rho=-0.36$ and $\rho=-0.37$) and this relationship strengthens for Log $\alpha$-Norm ($\rho=-0.55$ and $\rho=-0.40$).

Stable rank is seen to be positively associated with LiRA AUC. In HT-SR theory, stronger self-regularisation (characterised by lower stable rank and smaller $\alpha$) drives better generalisation. In this experimental setup, stable rank appears to function primarily as a proxy for model capacity (larger matrices naturally have higher stable rank). No correlation between stable rank and LiRA TPR@0.001 is observed on CIFAR-10 ($\rho=0.023$), however a moderate negative correlation is seen on Volkert ($\rho=-0.444$).

\begin{figure}
    \subfloat[CIFAR-10]{\includegraphics[width=0.24\linewidth]{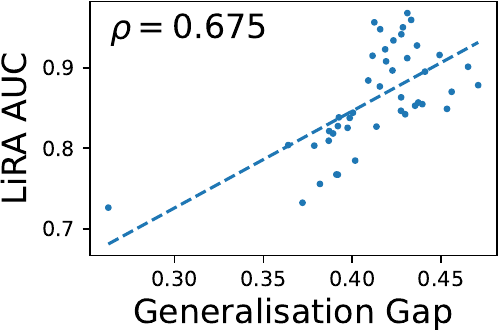}}
    \hfill
    \subfloat[CIFAR-10]{\includegraphics[width=0.24\linewidth]{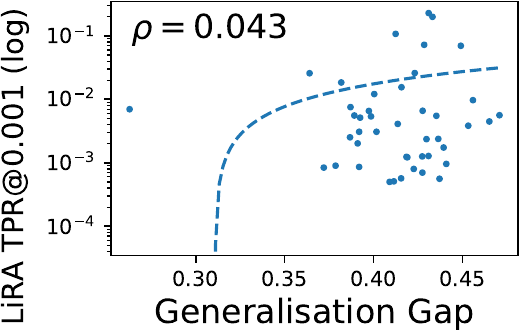}}
    \hfill
    \subfloat[CIFAR-10]{\includegraphics[width=0.24\linewidth]{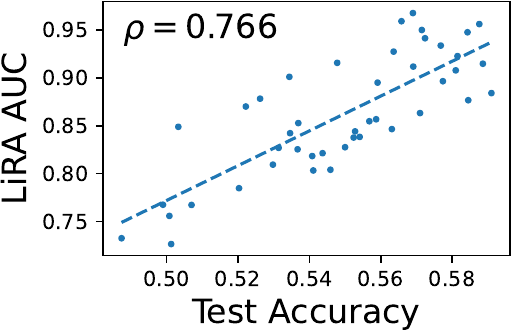}}
    \hfill
    \subfloat[CIFAR-10]{\includegraphics[width=0.24\linewidth]{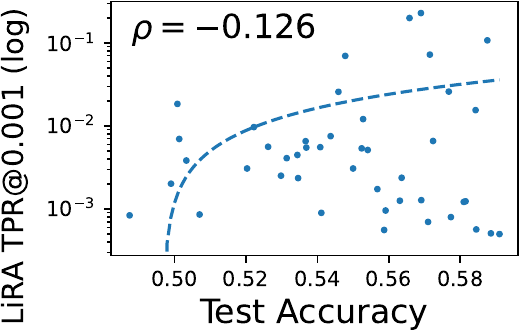}}\\
    \hfill

    \subfloat[Volkert]{\includegraphics[width=0.24\linewidth]{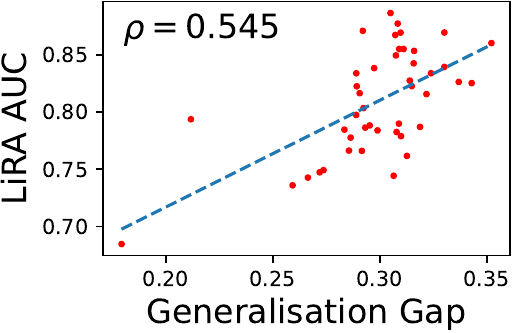}}
    \hfill
    \subfloat[Volkert]{\includegraphics[width=0.24\linewidth]{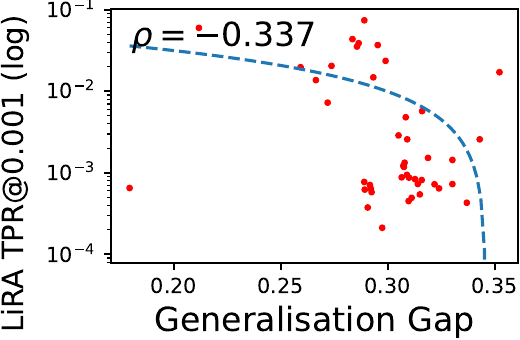}}
    \hfill
    \subfloat[Volkert]{\includegraphics[width=0.24\linewidth]{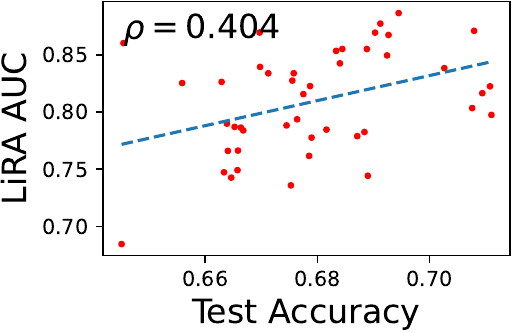}}
    \hfill
    \subfloat[Volkert]{\includegraphics[width=0.24\linewidth]{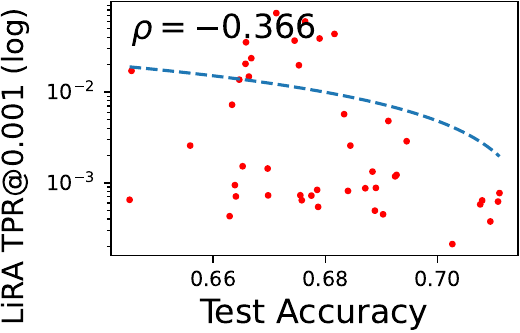}}\\
    \hfill

    \caption{Relationship between the generalisation metrics and privacy disclosure risk (LiRA AUC and TPR@0.001) on CIFAR-10 and Volkert datasets. LiRA TPR is shown on log-scale.}%
    \label{fig:ge_lira}
\end{figure}

\paragraph{Comparison with generalisation gap.}
Figure~\ref{fig:ge_lira} compares conventional measures of generalisation with LiRA privacy leakage. As expected, models with larger $\epsilon_{\text{gap}}$ are more vulnerable to MIAs, with positive correlations between $\epsilon_{\text{gap}}$ and LiRA AUC of $\rho=0.67$ on CIFAR-10 and $\rho=0.54$ on Volkert ($p\le0.01$), consistent with previous work~\citep{Yeom:2018, Carlini:2022}. However, these relationships are consistently weaker than those observed for stable rank. Similarly, the correlation between $\epsilon_{\text{gap}}$ and LiRA TPR@0.001 is weaker ($\rho=0.04$ on CIFAR-10 and $\rho=-0.34$ on Volkert) than the corresponding relationship with Log $\alpha$-Norm ($\rho=-0.55$ and $\rho=-0.40$). Since TPR@0.001 reflects attacker performance at an operationally relevant FPR, these results suggest that spectral metrics may provide stronger indicators of practically significant privacy leakage than the conventional $\epsilon_{\text{gap}}$.

This weaker relationship may partly reflect how $\epsilon_{\text{gap}}$ is constructed: within our hyperparameter grid, the gap appears to be driven predominantly by train accuracy rather than test accuracy. On CIFAR-10, $\epsilon_{\text{gap}}$ correlates strongly with train accuracy ($\rho=0.62$, $p<0.001$) but only weakly and non-significantly with test accuracy ($\rho=0.23$, $p=0.13$); the same asymmetry holds on Volkert (train accuracy: $\rho=0.47$, $p=0.001$; test accuracy: $\rho=-0.09$, $p=0.58$). This suggests that models in our grid diverge primarily through their degree of training-set memorisation rather than through differences in held-out generalisation performance, which may explain why $\epsilon_{\text{gap}}$ is a noisier predictor of LiRA vulnerability than stable rank: the gap conflates two components whose relationship to privacy leakage differ, whereas stable rank appears to track the memorisation-relevant component more directly.

\begin{figure}
    \subfloat[CIFAR-10]{\includegraphics[width=0.24\linewidth]{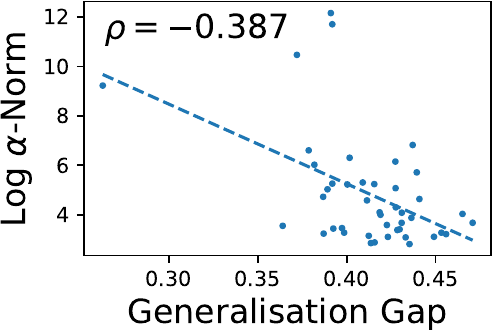}}
    \hfill
    \subfloat[CIFAR-10]{\includegraphics[width=0.24\linewidth]{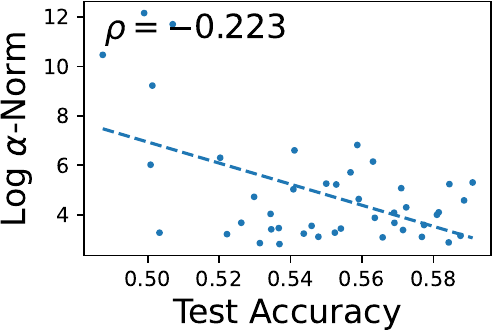}}
    \hfill
    \subfloat[CIFAR-10]{\includegraphics[width=0.24\linewidth]{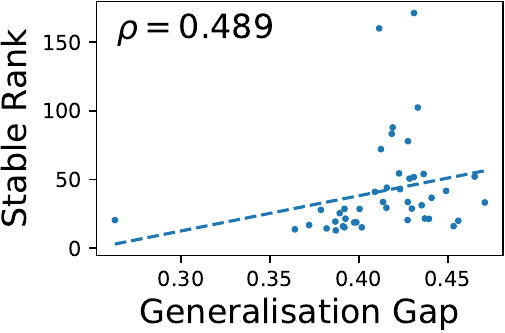}}
    \hfill
    \subfloat[CIFAR-10]{\includegraphics[width=0.24\linewidth]{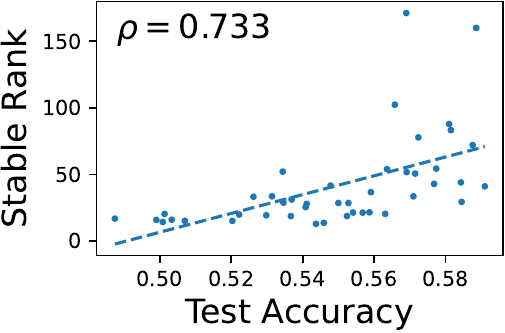}}\\
    \hfill

    \subfloat[Volkert]{\includegraphics[width=0.24\linewidth]{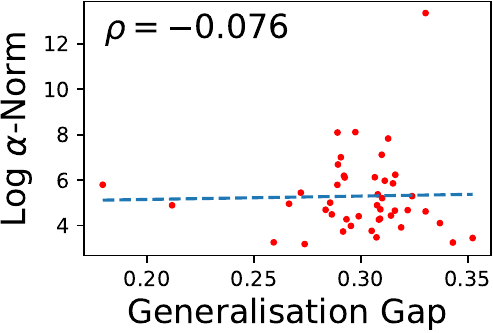}}
    \hfill
    \subfloat[Volkert]{\includegraphics[width=0.24\linewidth]{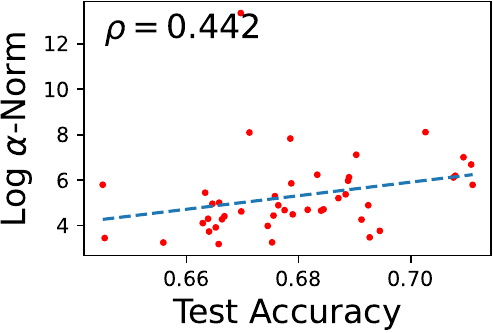}}
    \hfill
    \subfloat[Volkert]{\includegraphics[width=0.24\linewidth]{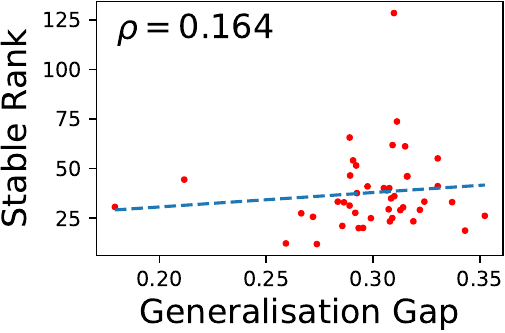}}
    \hfill
    \subfloat[Volkert]{\includegraphics[width=0.24\linewidth]{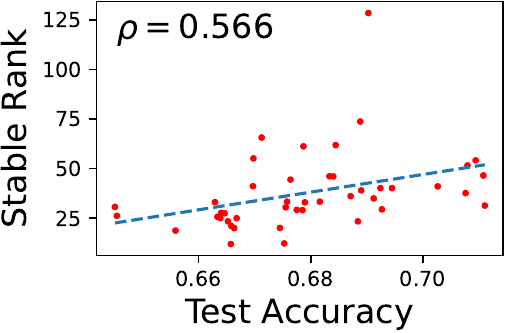}}\\
    \hfill

    \caption{Relationship between generalisation and spectral metrics (Log $\alpha$-Norm, and stable rank) on CIFAR-10 and Volkert datasets.}%
    \label{fig:ge_ww}
\end{figure}

\paragraph{Relationship between spectral metrics and generalisation.}
Figure~\ref{fig:ge_ww} investigates how spectral metrics relate to conventional measures of generalisation. Stable rank is positively correlated with test accuracy on both datasets ($\rho=0.73$ on CIFAR-10 and $\rho=0.57$ on Volkert; $p\le0.01$), indicating that the same spectral characteristics associated with improved predictive performance are also associated with increased privacy leakage. In contrast, Log $\alpha$-Norm exhibits only weak correlations with both test accuracy and $\epsilon_{\text{gap}}$, with the direction of these relationships differing between datasets. Together, these findings suggest that spectral metrics capture information beyond conventional generalisation measures, helping to explain their stronger association with LiRA privacy leakage.

A summary of correlations between spectral, generalisation, and privacy metrics is shown in Table~\ref{table:correlations}.

\begin{table}[t]
    \centering
    \caption{Summary of Spearman correlations ($\rho$) between spectral, generalisation, and privacy metrics.}
    \label{table:correlations}
    \begin{tabular}{lrrrr}
        \toprule
        & \multicolumn{2}{c}{\textbf{CIFAR-10}} & \multicolumn{2}{c}{\textbf{Volkert}} \\
        \cmidrule(lr){2-3} \cmidrule(lr){4-5}
        & \multicolumn{1}{c}{$\rho$} & \multicolumn{1}{c}{$p$-value} & \multicolumn{1}{c}{$\rho$} & \multicolumn{1}{c}{$p$-value} \\
        \midrule
        \textbf{WW metrics vs.\ LiRA AUC} & & & & \\
        $\alpha$ vs.\ AUC & $-0.08$ & $0.61$ & $0.61$ & $0.00$ \\
        Log $\alpha$-Norm vs.\ AUC & $-0.50$ & $0.00$ & $0.10$ & $0.52$ \\
        Log Spectral Norm vs.\ AUC & $-0.55$ & $0.00$ & $-0.27$ & $0.08$ \\
        Stable Rank vs.\ AUC & $0.87$ & $0.00$ & $0.60$ & $0.00$ \\
        \addlinespace
        \textbf{WW metrics vs.\ LiRA TPR@0.001} & & & & \\
        $\alpha$ vs.\ TPR@0.001 & $-0.36$ & $0.02$ & $-0.37$ & $0.01$ \\
        Log $\alpha$-Norm vs.\ TPR@0.001 & $-0.55$ & $0.00$ & $-0.41$ & $0.01$ \\
        Log Spectral Norm vs.\ TPR@0.001 & $-0.45$ & $0.00$ & $-0.18$ & $0.25$ \\
        Stable Rank vs.\ TPR@0.001 & $0.02$ & $0.88$ & $-0.44$ & $0.00$ \\
        \addlinespace
        \textbf{Generalisation metrics vs.\ Privacy} & & & & \\
        $\epsilon_{\text{gap}}$ vs.\ AUC & $0.68$ & $0.00$ & $0.54$ & $0.00$ \\
        $\epsilon_{\text{gap}}$ vs.\ TPR@0.001 & $0.04$ & $0.78$ & $-0.34$ & $0.03$ \\
        Test Acc.\ vs.\ AUC & $0.77$ & $0.00$ & $0.40$ & $0.01$ \\
        Test Acc.\ vs.\ TPR@0.001 & $-0.13$ & $0.41$ & $-0.37$ & $0.01$ \\
        \addlinespace
        \textbf{Generalisation metrics vs.\ WW metrics} & & & & \\
        $\epsilon_{\text{gap}}$ vs.\ Log $\alpha$-Norm & $-0.39$ & $0.01$ & $-0.08$ & $0.62$ \\
        $\epsilon_{\text{gap}}$ vs.\ Stable Rank & $0.49$ & $0.00$ & $0.16$ & $0.29$ \\
        Test Acc.\ vs.\ Log $\alpha$-Norm & $-0.22$ & $0.15$ & $0.44$ & $0.00$ \\
        Test Acc.\ vs.\ Stable Rank & $0.73$ & $0.00$ & $0.57$ & $0.00$ \\
        \addlinespace
        \textbf{Generalisation gap vs.\ Accuracy} & & & & \\
        $\epsilon_{\text{gap}}$ vs.\ Train Acc.\ & $0.62$ & $0.00$ & $0.47$ & $0.00$ \\
        $\epsilon_{\text{gap}}$ vs.\ Test Acc.\ & $0.23$ & $0.13$ & $-0.09$ & $0.58$ \\
        \bottomrule
    \end{tabular}
\end{table}

\paragraph{Combined predictors of privacy leakage.}

To assess whether stable rank and $\epsilon_{\text{gap}}$ capture complementary information about privacy leakage, we fit a multivariate ordinary least squares regression predicting LiRA AUC from both standardised predictors jointly. On CIFAR-10, the combined model explains substantially more variance ($R^2=0.71$) than either predictor alone ($\epsilon_{\text{gap}}$: $R^2=0.43$; stable rank: $R^2=0.47$), with both standardised coefficients remaining highly significant when included together (gap: $\beta=0.032$, $p<0.001$; stable rank: $\beta=0.035$, $p<0.001$). A similar pattern holds on Volkert, where the combined model again outperforms either individual predictor ($R^2=0.57$ vs.\ $0.38$ and $0.26$ respectively; gap: $\beta=0.026$, $p<0.001$; stable rank: $\beta=0.020$, $p<0.001$), although here $\epsilon_{\text{gap}}$ contributes a larger standardised effect than stable rank, reversing the ordering observed on CIFAR-10.

Using log stable rank in place of raw stable rank further increases performance on both datasets (combined $R^2=0.82$ on CIFAR-10 and $R^2=0.62$ on Volkert), and additionally resolves the coefficient-ordering reversal, with log stable rank contributing at least as much predictive weight as the generalisation gap on both datasets.

Together, these results indicate that stable rank and $\epsilon_{\text{gap}}$ provide complementary, non-redundant information about privacy leakage across both datasets.

We repeated this analysis for LiRA TPR@0.001, log-transforming the target to address strong right-skew in its raw distribution (consistent with the log-scale presentation in Figures~\ref{fig:lira_ww_cifar} and \ref{fig:lira_ww_volkert}). On CIFAR-10, the combined model explains modestly more variance than either predictor alone ($R^2=0.19$ vs.\ $0.16$ for Log $\alpha$-Norm and $0.001$ for $\epsilon_{\text{gap}}$ individually), with Log $\alpha$-Norm remaining the dominant, significant predictor ($\beta=-0.80$, $p=0.004$) while $\epsilon_{\text{gap}}$ does not reach significance ($\beta=-0.33$, $p=0.21$). On Volkert, the combined model again modestly outperforms either predictor alone ($R^2=0.16$ vs.\ $0.09$ and $0.07$ respectively), but here the pattern reverses: $\epsilon_{\text{gap}}$ is the (marginally) significant predictor ($\beta=-0.50$, $p=0.045$), while Log $\alpha$-Norm falls just short of significance ($\beta=-0.44$, $p=0.075$).

Combined explanatory power for TPR@0.001 is substantially lower than for AUC on both datasets, indicating that while spectral and generalisation-based metrics jointly explain most of the variance in overall attack success, predicting privacy leakage in the low-FPR regime remains considerably harder and the relative contribution of each predictor is less consistent across domains. Results are summarised in Table~\ref{table:ols}.

\begin{table}[t]
    \centering
    \caption{Ordinary least squares regression predicting LiRA AUC from standardised generalisation gap and log stable rank, and log LiRA TPR@0.001 from generalisation gap and Log $\alpha$-Norm ($n=44$ per dataset). $\beta$ denotes the standardised regression coefficient, reflecting the change in AUC (or log TPR) per one standard deviation change in the predictor.}
    \label{table:ols}
    \begin{tabular}{lrrrr}
        \toprule
        & \multicolumn{2}{c}{\textbf{CIFAR-10}} & \multicolumn{2}{c}{\textbf{Volkert}} \\
        \cmidrule(lr){2-3} \cmidrule(lr){4-5}
        & \multicolumn{1}{c}{Stat} & \multicolumn{1}{c}{$p$-value} & \multicolumn{1}{c}{Stat} & \multicolumn{1}{c}{$p$-value} \\
        \midrule
        \multicolumn{5}{l}{{\bf LiRA AUC}} \\
        $\epsilon_{\text{gap}}$ only ($R^2$) & 0.43 & & 0.38 & \\
        Log Stable Rank only ($R^2$) & 0.68 & & 0.32 & \\
        Combined $R^2$ & 0.82 & & 0.62 & \\
        \addlinespace
        $\beta_{\text{gap}}$ (combined) & 0.03 & 0.00 & 0.03 & 0.00 \\
        $\beta_{\text{log stable rank}}$ (combined) & 0.04 & 0.00 & 0.02 & 0.00 \\
        \addlinespace
        \multicolumn{5}{l}{{\bf (log) LiRA TPR@0.001}} \\
        $\epsilon_{\text{gap}}$ only ($R^2$) & 0.00 & & 0.09 & \\
        Log $\alpha$-Norm only ($R^2$) & 0.16 & & 0.07 & \\
        Combined $R^2$ & 0.19 & & 0.16 & \\
        \addlinespace
        $\beta_{\text{gap}}$ (combined) & $-0.33$ & 0.21 & $-0.50$ & 0.05 \\
        $\beta_{\text{Log }\alpha\text{-Norm}}$ (combined) & $-0.80$ & 0.00 & $-0.44$ & 0.08 \\
        \bottomrule
    \end{tabular}
\end{table}

\paragraph{Predicting high-risk models.}

To assess the practical utility for flagging high-risk models, we treat vulnerability as a binary outcome for each target model (LiRA TPR@0.001 $> 0.015$, corresponding to a $15\times$ improvement over random guessing) and compute ROC-AUC for each metric as a standalone classifier. Log $\alpha$-Norm and $\alpha$ achieve consistently strong separability on both CIFAR-10 (0.79, 0.75) and Volkert (0.70, 0.74). In contrast, $\epsilon_{\text{gap}}$ is only informative on Volkert (0.79) but performs near chance on CIFAR-10 (0.47), while stable rank shows the opposite pattern, performing reasonably on CIFAR-10 (0.67) but worse than random on Volkert (0.29).

This pattern is robust to the choice of threshold: repeating the analysis at $10\times$ and $20\times$ random guessing yields the same qualitative ordering, with $\alpha$ remaining the most consistently strong standalone predictor across both datasets (0.70--0.77) and $\epsilon_{\text{gap}}$ remaining weak on CIFAR-10 (0.40--0.50) but strong on Volkert (0.79--0.84) at every threshold tested.

Combining $\epsilon_{\text{gap}}$ with Log $\alpha$-Norm via logistic regression produces more consistent separability across datasets at every threshold tested (Table~\ref{table:threshold_robustness_combined}), leaving CIFAR-10 performance largely unchanged while substantially improving Volkert, most notably at $20\times$ where standalone Log $\alpha$-Norm performs much worse on Volkert than CIFAR-10 (0.61 vs.\ 0.83), but performance converges on the two datasets once combined (0.81 vs.\ 0.82). This suggests that combining spectral and generalisation-based signals offers a more dependable classifier than relying on any single metric alone, particularly as standalone metrics diverge across domains, consistent with the low-FPR regime being harder to predict (as observed above) and further motivating combined approaches for use in practice.

\begin{table}[t]
    \centering
    \caption{ROC-AUC for classifying models as vulnerable versus safe (LiRA TPR@0.001 above vs.\ below $10\times$, $15\times$, or $20\times$ random guessing) using Log $\alpha$-Norm and $\epsilon_{\text{gap}}$ individually, and combined via logistic regression, $n=44$ per dataset.}
    \label{table:threshold_robustness_combined}
    \begin{tabular}{lccc}
        \toprule
        & $10\times$ & $15\times$ & $20\times$ \\
        \midrule
        \multicolumn{4}{l}{\textbf{CIFAR-10}} \\
        Log $\alpha$-Norm & 0.74 & 0.79 & 0.83 \\
        $\epsilon_{\text{gap}}$ & 0.50 & 0.47 & 0.40 \\
        Combined & 0.73 & 0.77 & 0.82 \\
        \addlinespace
        \multicolumn{4}{l}{\textbf{Volkert}} \\
        Log $\alpha$-Norm & 0.70 & 0.70 & 0.61 \\
        $\epsilon_{\text{gap}}$ & 0.82 & 0.79 & 0.84 \\
        Combined & 0.81 & 0.77 & 0.81 \\
        \bottomrule
    \end{tabular}
\end{table}

\section{Conclusions and Limitations}%
\label{sec:conclusion}

\paragraph{Summary}

This paper presents a first investigation into whether inexpensive WW spectral metrics can be used to predict MIA privacy disclosure risk. Across image and tabular datasets, we find that stable rank is strongly associated with LiRA AUC, while Log $\alpha$-Norm exhibits a consistent negative association with LiRA TPR@0.001.

Importantly, these relationships are stronger than those observed using the conventional $\epsilon_{\text{gap}}$. Moreover, combining spectral and generalisation-based metrics improves prediction over either alone, indicating that spectral metrics and $\epsilon_{\text{gap}}$ capture different complementary aspects of privacy vulnerability.

To test whether spectral and generalisation-based metrics provide complementary rather than redundant information, we fit multivariate regressions combining $\epsilon_{\text{gap}}$ with the strongest associated WW metric for each privacy outcome. Combining $\epsilon_{\text{gap}}$ and (log) stable rank substantially improved prediction of LiRA AUC over either predictor alone on both datasets, with both coefficients remaining significant jointly, indicating genuinely complementary signal. Gains for LiRA TPR@0.001 were more modest, and which predictor dominated reversed between datasets, suggesting that the low FPR regime is inherently harder to predict consistently across domains.

Beyond continuous prediction, we further show that treating vulnerability as a binary outcome enables high-risk models to be flagged: $\alpha$ and Log $\alpha$-Norm reliably separate high-risk from low-risk models on both datasets, whereas $\epsilon_{\text{gap}}$ and stable rank do not generalise consistently across domains for this task. Combining $\epsilon_{\text{gap}}$ with Log $\alpha$-Norm offers a modest, more consistent improvement over either metric alone, echoing the difficulty of predicting leakage in the low-FPR regime observed in the continuous analysis.

These findings indicate that spectral properties of neural network weight matrices provide a promising and computationally efficient signal for assessing privacy risk without requiring expensive attack simulations. Since spectral metrics can be computed directly from trained model weights, they offer the potential for scalable privacy auditing and model selection based on privacy characteristics.

While the results demonstrate consistent correlations across two datasets, they do not establish a causal relationship between spectral properties and privacy leakage. Our working hypothesis assumes that WW metrics predict MIA vulnerability indirectly by acting as proxies for model generalisation and overfitting, with generalisation itself serving as a proxy for privacy risk. However, the stronger associations observed between WW metrics and LiRA performance than between $\epsilon_{\text{gap}}$ and LiRA suggest that this hypothesis may be incomplete. Future work should therefore investigate whether spectral metrics directly characterise properties of the loss distributions exploited by likelihood-based attacks. In addition, this study considers only pairwise combinations and monotonic relationships; richer multivariate or non-linear combinations of spectral features, potentially incorporating all WW metrics jointly, may provide substantially stronger predictors of privacy disclosure risk, particularly in the harder to predict low FPR regime.

\paragraph{Limitations.}

This study is limited to a single image dataset (CIFAR-10) and a single tabular dataset (Volkert) with MLPs trained for 100 epochs. Although these datasets provide complementary domains and CIFAR-10 is one of the most widely used benchmarks in MIA research~\citep{Hu:2022}, it remains unclear whether the observed relationships extend to larger, more complex architectures (e.g., ResNets and Transformers) or across more diverse data modalities (e.g., text). Additionally, we evaluate only one MIA method (LiRA). While LiRA is widely regarded as a strong MIA, alternative attack formulations may exhibit different relationships with spectral properties. Our multivariate analysis is similarly limited in scope: with only 44 models per dataset, the regressions should be interpreted as suggestive rather than definitive, and we did not attempt cross-dataset transfer or held-out validation of the fitted models, which would provide a more stringent test of generalisability. Consequently, the results should be interpreted as evidence that WW metrics are promising indicators of LiRA privacy risk, rather than as establishing their general applicability across architectures, datasets, or attack methodologies.

\begin{ack}
This work was funded by UK Research and Innovation (Grant Number MC\_PC\_24038) as part of the Data and Analytics Research Environments UK (DARE UK) programme, delivered in partnership with Health Data Research UK (HDR UK) and Administrative Data Research UK (ADR UK). The authors wish to thank Charles H. Martin for discussions of WeightWatcher.
\end{ack}

%

\bibliographystyle{plainnat}
\bibliography{references}

\end{document}